\pdfoutput=1

\documentclass[11pt]{article}

\usepackage[final]{acl}

\usepackage{times}
\usepackage{latexsym}
\usepackage{amsmath}
\usepackage{verbatim}

\usepackage[T1]{fontenc}

\usepackage[utf8]{inputenc}

\usepackage{microtype}

\usepackage{inconsolata}

\usepackage{color, colortbl}
\usepackage{graphicx}
\usepackage{url}            
\usepackage{booktabs}       
\usepackage{amsfonts}       
\usepackage{xcolor}         
\usepackage{makecell}
\usepackage{arydshln}
\usepackage{wrapfig}
\usepackage{multirow}

\definecolor{lightblue}{rgb}{0.93,0.95,1.0} 

\usepackage{algorithm}
\usepackage{algpseudocode}

\usepackage{amssymb}

\title{MMS-LLaMA: Efficient LLM-based Audio-Visual \\ Speech Recognition with Minimal Multimodal Speech Tokens}

\author{Jeong Hun Yeo$^*$, Hyeongseop Rha$^*$, Se Jin Park, Yong Man Ro$^\dagger$
\\Integrated Vision and Language Lab, KAIST, South Korea\\
\small{\texttt{\{sedne246, ryool\_1832, jinny960812, ymro\}@kaist.ac.kr}}}

\begin{document}
\maketitle
\def\thefootnote{}\footnotetext{$^*$Equal Contribution. $^\dagger$Corresponding Author.}
\begin{abstract}
Audio-Visual Speech Recognition (AVSR) achieves robust speech recognition in noisy environments by combining auditory and visual information. However, recent Large Language Model (LLM) based AVSR systems incur high computational costs due to the high temporal resolution of audio-visual speech processed by LLMs. In this work, we introduce an efficient multimodal speech LLM framework that minimizes token length while preserving essential linguistic content. Our approach employs an early AV-fusion module for streamlined feature integration, an audio-visual speech Q-Former that dynamically allocates tokens based on input duration, and a refined query allocation strategy with a speech rate predictor to adjust token allocation according to speaking speed of each audio sample. Extensive experiments on the LRS3 dataset show that our method achieves state-of-the-art performance with a WER of 0.72\% while using only 3.5 tokens per second. Moreover, our approach not only reduces token usage by 86\% compared to the previous multimodal speech LLM framework, but also improves computational efficiency by reducing FLOPs by 35.7\%. The code and models are available \url{https://github.com/JeongHun0716/MMS-LLaMA}.
\end{abstract}

\section{Introduction}
In human communication, watching lip movements and listening to sounds are essential for understanding speech. These multimodal cues, which combine visual and auditory information, enable people to communicate effectively in bustling cafes, crowded streets, and noisy factories. Thanks to these practical advantages and significant advances in deep learning, Audio-Visual Speech Recognition (AVSR) technology has made remarkable progress through numerous research efforts  \cite{shi2022learning, hu2022interactive, ma2023auto, afouras2018deep, ma2021end, serdyuk2022transformer, fu2024boosting, cappellazzo2024large}. Now, it is easy to find an AVSR model that can accurately predict what you have said, even in noisy environments.

Rapid progress has been driven by large-scale audio-visual datasets \cite{afouras2018lrs3, afouras2018deep}, advanced neural architectures \cite{elman1990finding, vaswani2017attention, gulati2020conformer}, improved multimodal learning strategies, including self-supervised learning \cite{shi2022learning} and knowledge distillation using a pre-trained Automatic Speech Recognition (ASR) model \cite{ma2023auto}; carefully designed training methods \cite{ma2022visual,hong2023watch}; and the utilization of Large Language Models' (LLMs) language understanding capabilities as sentence predictors \cite{cappellazzo2024large}. Among these, multimodal speech LLM frameworks stand out by integrating LLMs for enhanced context modeling. However, despite these gain, they still suffer from high computational costs. This is largely due to multimodal speech tokens having a higher temporal resolution compared to text tokens, which forces the self-attention mechanism in each LLM layer to process many more tokens, thereby significantly increasing the computational burden.

To address these issues, we aim to develop an efficient MultiModal Speech LLM framework (MMS-LLaMA) that minimizes the lenght of multimodal speech tokens while preserving their linguistic content. To achieve this, we construct this framework with three primary components: 1) early AV-fusion module shifts the fusion process to an earlier stage, prior to inputting multimodal speech tokens into the LLM. 2) Audio-Visual speech Q-former (AV Q-Former) is designed to dynamically allocate the number of learnable queries according to the duration of audio-visual input, where queries are transformed into multimodal speech tokens. 3) Going one step further, to more effectively adjust the number of multimodal tokens, we propose a speech rate predictor. It accounts for differences in speaking speed across each audio sample, enabling the assignment of more tokens to faster speech.

Reflecting the growing trend toward efficient multimodal models and leveraging insights from efficient multimodal LLMs that utilize minimal vision tokens during pre-fusion \cite{zhang2025llava}, we incorporate their effective pre-fusion approach into our multimodal speech LLM framework. Concurrently, we explore and integrate existing audio-visual speech fusion strategies into our early-AV fusion module, which reduces the number of fused sequences by half.

Even though the number of fused sequences is reduced by half, their temporal resolution remains higher than that of text tokens. To minimize the audio-visual sequence length to similar level of text, AV Q-Former is designed. Since the language content in audio-visual inputs scales with their duration, a fixed-length output from a conventional Q-Former \cite{dai2023instructblip} may be inefficient for compressing variable-length audio and visual modalities. To address this, we propose a novel query allocation strategy that adjusts the number of multimodal speech tokens based on the duration of the audio-visual input. Using this dynamic strategy, we empirically determine the minimum number of multimodal speech tokens required for variable-length inputs and compress the sequence to a scale comparable to that of text tokens. Finally, to further enhance efficiency, we refine our query allocation strategy by incorporating a speech rate predictor. Even if two audio-visual inputs have the same duration, their language content may differ depending on the speaker's speaking rate. By taking this into account, we can assign more queries to faster speech segments, thereby generating a greater number of multimodal speech tokens and preserving their linguistic content. 

Through extensive experiments, we demonstrate that using only 3.5 multimodal speech tokens per second can effectively preserve language content while maintaining performance. Moreover, our approach achieves state-of-the-art performance with a Word Error Rate (WER) of 0.72\% on the LRS3 \cite{afouras2018lrs3} dataset. Compared to the previous LLM-based AVSR model \cite{cappellazzo2024large} that uses 25 tokens per second, our method reduces token usage by 86\%. Additionally, our approach improves computational efficiency, reducing FLOPs by 35.7\%.

\section{Related Work}
\subsection{Audio-Visual Speech Recognition}
Audio-based automatic speech recognition is a well-studied and one of the most frequently used technology \cite{amodei2016deep,kim2017joint,prabhavalkar2023end}. However, as it primarily depends on acoustic inputs, its performance naturally degraded when the input audio is perturbed with other background noises (e.g., babble noise) \cite{varga1993assessment}. Visual Speech Recognition (VSR) aims to employ visual inputs only in speech recognition, using the fact that the visual inputs are not affected by the acoustic noises. A diverse range of literature has been explored on VSR techniques to avoid the impact of background noise on speech recognition performance \cite{petridis2017end,petridis2018end,martinez2020lipreading,ma2021end,ma2022visual,ma2023auto,kim2022distinguishing, cheng2023opensr, kim2023lip,kim2024prompt,kim2024efficient,yeo2024personalized,yeo2024akvsr,yeo2024visual2}.

Audio-visual speech recognition integrates both audio and visual modalities as input, combining the strengths of ASR and VSR techniques to enhance the robustness and performance of speech recognition systems, particularly in noisy environments. Early AVSR models \cite{noda2015audio, huang2013audio, mroueh2015deep, stewart2013robust} established the foundation for multimodal speech recognition, demonstrating the effectiveness of fusing visual features such as lip movements with audio signals. Advancements in AVSR \cite{petridis2018audio,shi2022robust,hong2022visual,hong2023watch} have been fueled by improvements in both data availability \cite{chung2017lip,afouras2018lrs3} and model architectures \cite{vaswani2017attention,gulati2020conformer}. Self-supervised learning has also played a crucial role in advancing AVSR \cite{shi2022learning, lian2023av, haliassos2024braven, haliassos2024unified} to further enhance performance.

More recently, two research trends have emerged in the field. One focuses on leveraging pre-trained ASR models, with approaches like pseudo-labeling unlabeled audio-visual data for data augmentation \cite{ma2023auto,yeo2024visual2} or integrating a pre-trained visual encoder with Whisper \cite{rouditchenko2024whisper}. The other trend integrates LLMs with speech features to harness the context modeling capabilities of LLMs, thereby enhancing the recognition performance \cite{yu2024connecting,yeo2024visual,cappellazzo2024large}.

While recent efforts leveraging LLMs have achieved remarkable performance in speech recognition, they have primarily focused on further improving accuracy. In contrast, our goal is to limit the computational burden on our multimodal speech LLM without sacrificing accuracy, we introduce the AV Q-Former. By dynamically modifying the number of multimodal speech tokens, it effectively preserves essential linguistic details from both the audio and visual streams.

\begin{figure*}[t]
\centering
\centerline{\includegraphics[width=15.5cm]{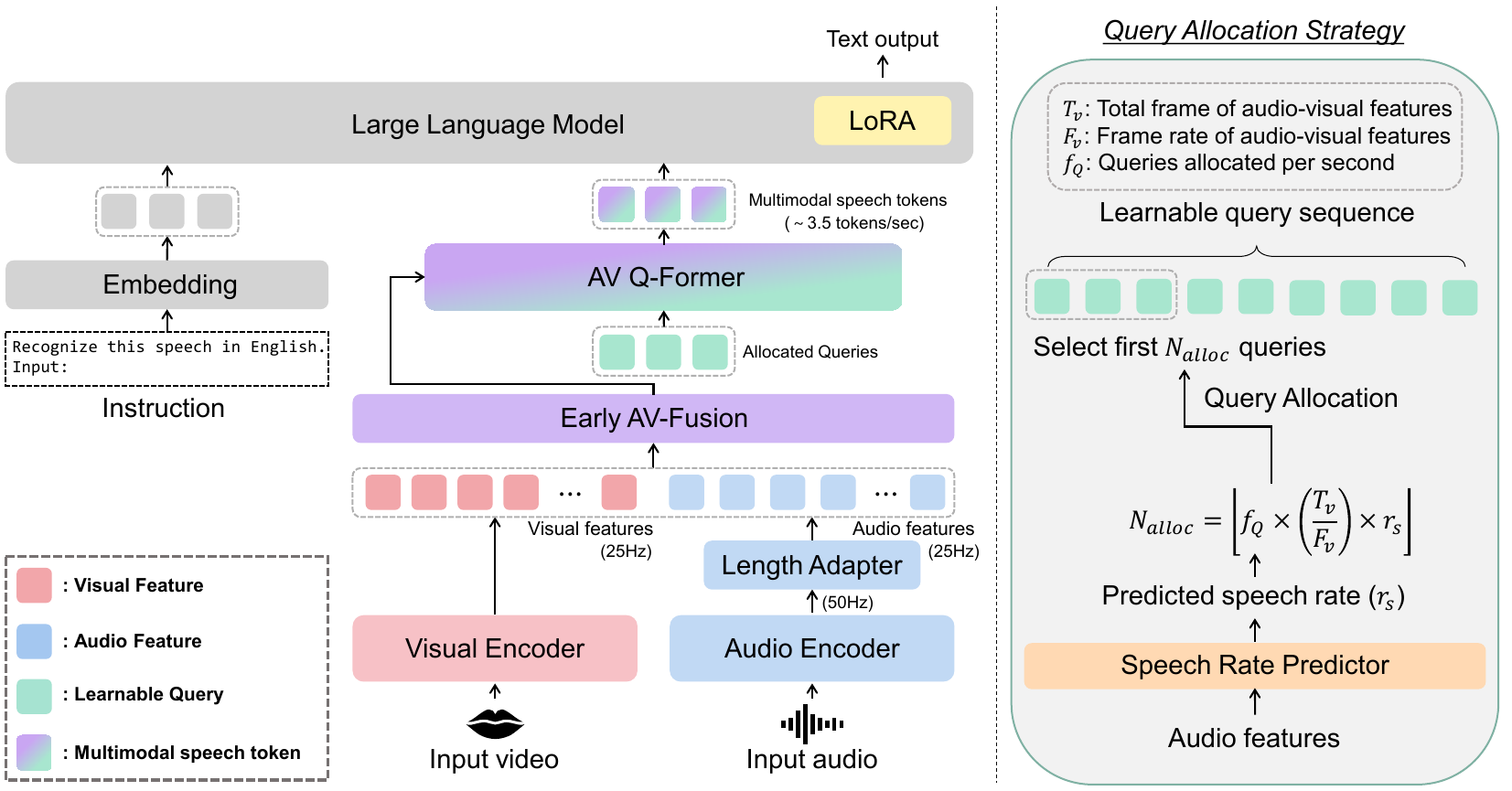}}
\caption{Illustration of the MMS-LLaMA framework. Audio and visual features are extracted by separate encoders, with a length adapter aligning audio frames. Early AV-Fusion merges the two modalities, while a speech rate predictor estimates speaking speed. Through the query allocation strategy, the appropriate number of queries is passed to the AV Q-Former, which produces multi-modal speech tokens. These tokens, combined with instruction embeddings, are then fed into the LLM to generate the output sentence. During training, the parameters of the encoders, the LLM’s text embedding layer, and the speech rate predictor remain frozen.}
\label{fig:1}
\end{figure*}

\subsection{Speech Large Language Model}
Advancements in speech recognition have been achieved through self-supervised learning \cite{hsu2021hubert} and the development of large-scale ASR datasets \cite{radford2023robust}. Building on these models, more recent speech recognition systems \cite{fathullah2024prompting, rubenstein2023audiopalm} have significantly improved performance, especially for low-resource languages, by leveraging the multilingual capabilities of LLMs. Additionally, by utilizing LLM's capabilities, error correction methods \cite{chen2023hyporadise, chen2024s, radhakrishnan2023whispering} have further enhanced the reliability and accuracy of ASR outputs. Beyond speech recognition, Speech LLMs have also expanded to support multitask learning, enabling a single model to perform a wide range of speech-related tasks \cite{chu2023qwen, tang2023salmonn}. Qwen-Audio \cite{chu2023qwen} and SALMONN \cite{tang2023salmonn} scale audio-language pre-training to cover various speech-related tasks and diverse audio inputs. 


Despite this significant progress, the exploration of extending LLMs' capabilities to the audio-visual speech domain remains limited. Moreover, most current speech LLMs primarily focus on audio-based tasks and have under-explored the effectiveness of adopting the utilization of speech rate. In contrast, our work incorporates a speech rate predictor that dynamically allocates resources based on the speaking speed of each audio, enabling more efficient and robust processing for audio-visual speech inputs.

\section{Method}
In this paper, our objective is to minimize the length of multimodal tokens while preserving their linguistic content, thereby enhancing the efficiency of our multimodal speech LLM framework. Recent works \cite{yeo2024visual,cappellazzo2024large} have demonstrated that LLMs can serve as effective multimodal speech learners by leveraging their context modeling capabilities. Motivated by this finding, we adopt their framework as our baseline AVSR model.

As illustrated in Figure~\ref{fig:1}, the architecture consists of three main components: a visual encoder, an audio encoder, and an LLM decoder that predicts sentences from multimodal tokens. Building on these components, and with the goal of reducing the number of multimodal tokens while retaining essential linguistic information, we introduce three additional modules: the early AV-fusion module, the AV Q-Former, and the speech rate predictor.

\subsection{Early AV-fusion Module}
Although pre-fusion techniques that combine visual and text modalities have proven effective at reducing computational costs, their application in multimodal speech LLM frameworks remains unexplored. To extend their effectiveness to the multimodal speech domain, we propose an early AV-fusion module that fuses visual and audio modalities before inputting them into the LLM, thereby halving the sequence length. To design this module effectively, we investigate three previously proposed fusion techniques for audio-visual speech: concatenation, addition, and multimodal attention.

Given the audio and video inputs, the visual encoder and audio encoder extract visual features $\mathbf{X}_{v} \in \mathbb{R}^{T_{v} \times D} $ and audio features $\mathbf{X}_{a} \in \mathbb{R}^{T_{a} \times D} $ that contain linguistic content from lip movements and sound, respectively. Since these features have different temporal resolutions (with audio features typically having a higher resolution than visual features), we employ a length adapter to resample the audio features so that they match the temporal scale of the visual features. We denote the resampled audio features as $\mathbf{X}'_{a} \in \mathbb{R}^{T_{v} \times D} $. Through this process, we align the audio-visual features along the time dimension and evaluate the effectiveness of three fusion methods based on these aligned features.

\textbf{Concatenation.}
The audio and visual feature vectors are combined by simply appending one to the other along the feature dimension via concatenation approach. This can be expressed as follows:
\begin{equation}
    \mathbf{X}_{av} = [\mathbf{X}'_a;\mathbf{X}_v] \in \mathbb{R}^{T_v \times 2D}
\end{equation}

\textbf{Addition.}
The addition method fuses audio and visual features by performing an element-wise sum.  It can be formulated as follows: 
\begin{equation}
    \mathbf{X}_{av} = \mathbf{X}'_a + \mathbf{X}_v \in \mathbb{R}^{T_v \times D}
\end{equation}
where, $+$ indicate element-wise summation. 

\textbf{Multimodal Attention.}
The multimodal attention method fuses audio and visual features based on attention mechanism.
\begin{equation}
    \mathbf{X}_{av} = \operatorname{MHCA}(\mathbf{X}_vW_Q, \mathbf{X}'_aW_K, \mathbf{X}'_aW_V) 
\end{equation}
where $\operatorname{MHCA}$ indicate multi-head cross attention, $\mathbf{X}_{av} \in \mathbb{R}^{T_v \times D}$, $W_Q$, $W_K$, and $W_V$  are learnable projection matrices that transform the features into the query, key, and value.

\subsection{AV Q-Former}
While these early AV-fusion modules reduce the length of the audio-visual feature sequence by half, there is still a gap compared to the number of text tokens. To bridge this gap between audio-visual speech and text modalities in terms of token count, we introduce a novel AV Q-Former.

To transform variable-length input sequences into fixed-length output queries, the Q-Former is introduced by \citet{dai2023instructblip} in the vision-language domain. By employing a fixed-size window, \citet{tang2023salmonn} demonstrates that the Q-Former effectively compresses audio-based speech tokens into text-level queries. Despite this progress, because the window-level Q-Former uses a fixed window size, it captures context information from only a portion of speech tokens in a single query. To address this limitation, we employ the conventional Q-Former with a novel query allocation strategy.

\subsubsection{Query Allocation Strategy}
While Q-Formers allocate a fixed number of queries regardless of input sequence length, the language content of audio-visual inputs is proportional to their duration. Therefore, our allocation strategy aims to dynamically adjust the number of queries based on the input length.

As shown in Figure~\ref{fig:1}, the AV Q-Former dynamically assigns a number of queries proportional to the length of the audio-visual feature sequence. To achieve this, we define a learnable query sequence  $\mathbf{Q} \in \mathbb{R}^{N \times D_{q}}$, where $N$ denotes the number of queries and $D_{q}$ represents the embedding dimension of each query. Then, depending on the duration of the input audio and video, the number of queries is allocated proportionally to their respective durations. Let $F_v$ denote the frame rate (in Hz) of the audio-visual feature sequence and assume a query rate $f_Q$ (i.e., the number of queries per second), the allocated number of queries is given by

\begin{equation}
    N_{alloc} = \lfloor f_Q \times \frac{T_{v}}{F_{v}} \rfloor
\end{equation}

We subsequently select the first $N_{alloc}$ queries from the learnable query sequence. It can be formulated as: $\mathbf{Q}_{alloc}=\mathbf{Q}[:N_{alloc}] \in \mathbb{R}^{N_{alloc} \times D_{q}}$. These queries with audio-visual feature sequence, are fed into the Q-Former to generate multimodal speech tokens $\mathbf{M} \in \mathbb{R}^{N_{alloc} \times D_{q}}$. This process can be expressed as follows:

\begin{equation}
    \mathbf{M} = \text{Q-Former}(\mathbf{Q}_{\text{alloc}}; \mathbf{X}_{\text{av}})
\end{equation}

Then, we apply two linear layers to project the multimodal speech tokens into the LLM's embedding space. Next, we concatenate these projected tokens with the text instruction embeddings along the temporal axis, and then provide the resulting sequence as input to the LLM to predict the sentence. With this allocation strategy, we explore the minimum query frequency required to effectively compress the audio-visual feature sequence while maintaining performance.

\begin{figure}[t]
\centering
\centerline{\includegraphics[width=7.5cm]{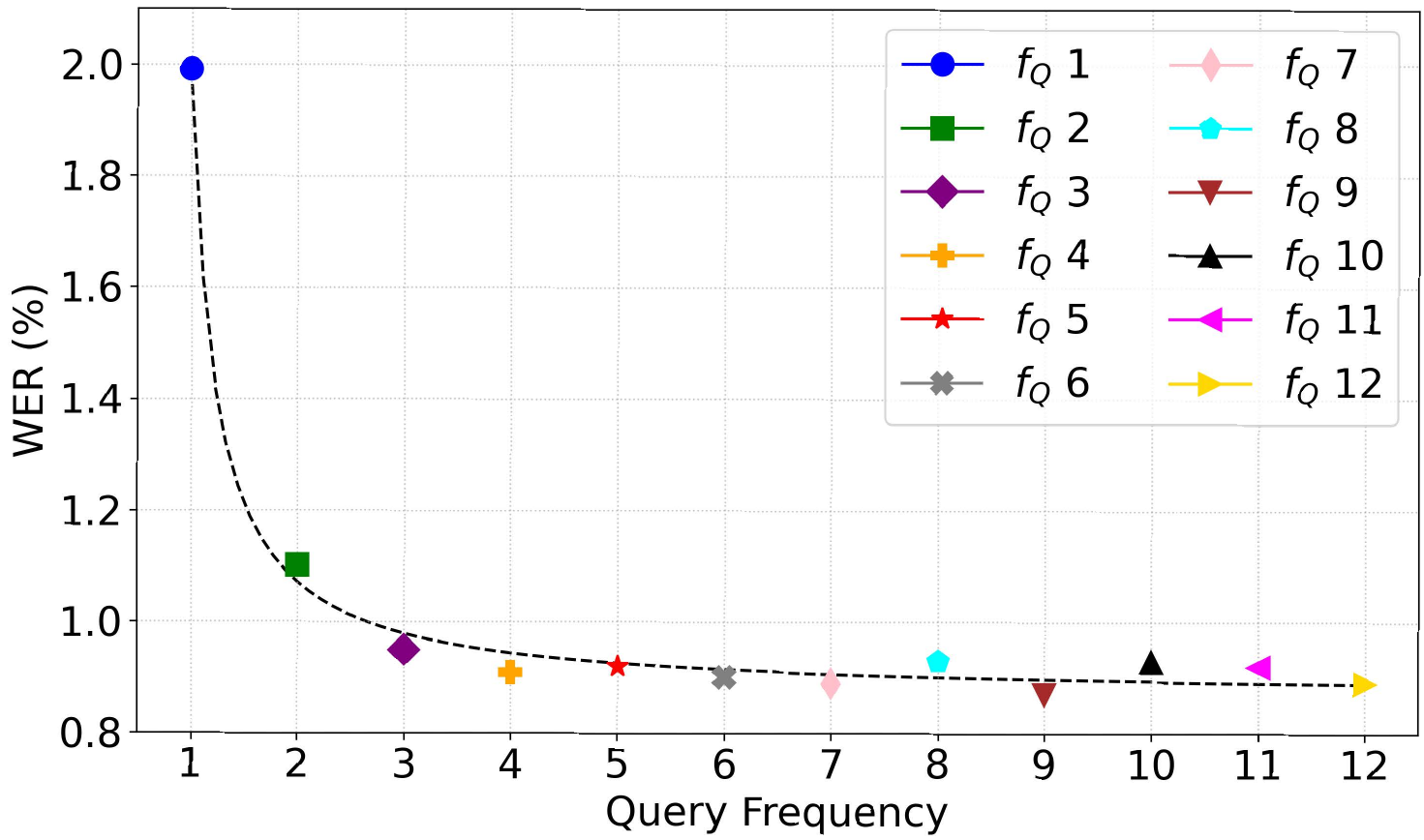}}
\caption{Impact of Query Frequency Variations on AVSR Performance. This results demonstrates that compressing the audio-visual feature sequence to as low as query frequency 4 maintains performance.}
\label{fig:2}
\end{figure}
\subsection{Speech Rate Predictor}
Through our AV Q-Former, as shown in Figure~\ref{fig:2}, we have confirmed that that performance remains robust even when compressing the audio-visual feature sequence by leveraging a query frequency of 4. However, performance begins to degrade below a query frequency of 4 Hz. This is likely due to variations in the speech rate across audio samples, which rate usually measured in words per minute. Faster speech may contain more linguistic content and thus require additional queries, even if the total duration is the same as slower speech. To address this, we propose a speech rate predictor that allocates queries more effectively by considering each audio sample's speech rate. Our goal with this predictor is to optimize the query allocation strategy for enhanced efficiency. 

To train the speech rate predictor, we first compute the average speech rate across the training set and normalize each sample’s speech rate based on this reference. The normalized values serve as target labels, and we train the predictor using Mean Squared Error (MSE) loss with only audio features as input. This training process is performed before training our multimodal speech LLM framework. The pre-trained predictor then estimates the speech rate $r_s$ and allocates more queries to higher speech rates (i.e., faster speech). This process can be formulated as follows:
\begin{equation}
    N_{alloc} = \lfloor f_Q \times \frac{T_{v}}{F_{v}} \times r_s \rfloor.
\end{equation}
The illustration of the speech rate predictor can be found in the right side of Figure~\ref{fig:1}.

\begin{table*}[t]
  \renewcommand{\arraystretch}{1.5}
  \renewcommand{\tabcolsep}{2.5mm}
  \centering
  \resizebox{0.95\linewidth}{!}{
  \begin{tabular}{ccccccc}
    \Xhline{3\arrayrulewidth}
    \multirow{2}{*}{\textbf{Method}} 
    & \multirow{2}{*}{\makecell{\textbf{Audio} \\ \textbf{Encoder}}} 
    & \multirow{2}{*}{\makecell{\textbf{Visual} \\ \textbf{Encoder}}}
    & \multirow{2}{*}{\makecell{\textbf{Decoder}}}
    & \multirow{2}{*}{\makecell{\textbf{Training} \\ \textbf{Data(h)}}}
    & \multicolumn{2}{c}{\textbf{WER(\%)$\downarrow$}}  \\
    \cline{6-7}
    &  & & & & \textbf{Noisy} & \textbf{Clean} \\
    \hline
    CM-seq2seq \cite{ma2021end} & \multicolumn{2}{c}{Conformer} & Transformer & 433 & - & 2.3 \\
    ViT3D-CM \cite{serdyuk2022transformer} & \multicolumn{2}{c}{Conformer} & LSTM & 90K  & 2.9 & 1.6 \\
    CMA \cite{kim2024learning} & \multicolumn{2}{c}{Transformer} & Transformer & 433 &  4.4 & 1.5 \\
    AV-data2vec \cite{lian2023av} & \multicolumn{2}{c}{Transformer} & Transformer & 433  & 6.7 & 2.5 \\
    auto-avsr \cite{ma2023auto} & \multicolumn{2}{c}{Conformer} & Transformer & 1902/3448  &  - & 1.0/0.9 \\
    LP Conformer \cite{chang2024conformer} & \multicolumn{2}{c}{Conformer} & LSTM & 100K  & \textbf{1.9} & 0.9 \\
    Whisper-Flamingo \cite{rouditchenko2024whisper} & Whisper & AV-HuBERT & Whisper & 433/1759  &  5.6/5.6 & 1.1/0.76 \\
    \hline
    \multicolumn{7}{>{\columncolor{gray!20}}c}{\textbf{Multi-modal Speech LLM Framework}} \\
    \multirow{2}{*}{LLaMA-AVSR \cite{cappellazzo2024large}}
     & Whisper & AV-HuBERT & LLaMA 3.1 8B & 433  & 4.2 & 0.95 \\
     & Whisper & AV-HuBERT & LLaMA 3.1 8B & 1759 & - & 0.77 \\
    \hdashline
    \multirow{2}{*}{\textbf{MMS-LLaMA}}
     & Whisper & AV-HuBERT & LLaMA 3.2 3B & 433 & 2.4 & 0.9 \\
     & Whisper & AV-HuBERT & LLaMA 3.2 3B & 1759 & \textbf{1.9} & \textbf{0.72} \\
    \hline
  \end{tabular}}
  \caption{Comparisons with state-of-the-art methods on the LRS3 dataset. We report each method’s architecture (audio encoder, visual encoder, decoder), the amount of training data, and WER under both clean and noisy conditions. The clean condition is evaluated on the original test set, while the noisy condition is evaluated on a test set with babble noise added at 0-SNR}
  \label{table:1}
\end{table*}

\section{Experimental Setup}
\subsection{Dataset}
\noindent{\bf Lip Reading Sentences 3 (LRS3)}, as detailed in \cite{afouras2018lrs3}, is a widely used dataset designed for AVSR. It includes 433 hours of audio-visual data for training and 1 hour for evaluation. The videos are sourced from TED and TEDx talks and are accompanied by human-annotated text transcriptions.

\noindent{\bf VoxCeleb2} as detailed in \cite{chung2018voxceleb2}, is a dataset designed for speaker recognition. It consists of 2,442 hours of multilingual audio-visual data. Following \cite{shi2022learning}, we utilize only the English portion of this dataset, which amounts to 1,326 hours. Moreover, we also use the Whisper ASR model to generate pseudo text transcriptions, which we combine with the LRS3 dataset for training our model. This combined dataset amounts to 1,759 hours.

\subsection{Implementation Details}
\subsubsection{Pre-processing}
Following \cite{ma2023auto}, we resample all audio and video from the LRS3 and VoxCeleb2 datasets to 25 fps and 16 kHz, respectively. Using RetinaFace \cite{deng2020retinaface}, we crop the mouth region from the face video to a size of 96x96. The cropped mouth clips are then converted to grayscale and flipped horizontally for data augmentation during the training stage. Audio at a 16 kHz sampling rate is mixed with babble noise from the NOISEX dataset \cite{varga1993assessment} at a 75\% probability. After Whisper processing, the audio is padded to 30 seconds and converted into an 80-dimensional Mel spectrogram, which is then fed into the Whisper encoder

\subsubsection{Architectures}
We adopt AV-HuBERT \cite{shi2022learning} as the visual encoder and Whisper \cite{radford2023robust} as the audio encoder. For the large language model (LLM), we use LLaMA variants: LLaMA 3.2 1B, 3B, and LLaMA 3.1 8B \cite{dubey2024llama}. Our AV Q-Former is based on a BERT-large model with two Transformer layers, each having an embedding dimension of 1024, 16 attention heads, and a feed-forward dimension of 4096. Finally, the speech rate predictor consists of two Transformer layers with a 256-dimensional embedding, 4 attention heads, and a feed-forward dimension of 1024.

\subsubsection{Training and evaluation}
We use the Adam optimizer with $\beta_1=0.9$ and $\beta_2=0.98$, alongside a cosine learning rate scheduler. The initial learning rate is set to $1e^{-4}$, with 0.5k warm-up steps out of a total 30,000 steps. We also employ a minimum learning rate of $1e^{-5}$ and a final learning rate scale of 0.05. For fine-tuning the LLM, we adopt the QLoRA \cite{dettmers2024qlora} approach with a LoRA rank of 16, an alpha (scaling factor) of 32, and a dropout rate of 0.05, applying LoRA to the query, key, value, and output projection layers. We evaluate performance using beam search decoding (beam size = 5) with a temperature of 0.3. All experiments are conducted on 8 RTX 3090 GPUs.

\section{Experimental Results}
\subsection{Comparison with the state-of-the-art methods}
In order to validate the effectiveness of the proposed method, we compare the proposed MMS-LLaMA with the previous state-of-the-art AVSR methods on LRS3 dataset, as shown in Table~\ref{table:1}.

Traditional approaches, such as CM-seq2seq~\cite{ma2021end}, ViT3D-CM~\cite{serdyuk2022transformer}, and auto-avsr~\cite{ma2023auto}, commonly employ Conformer- or Transformer-based backbones and rely on large-scale datasets to improve AVSR performances. Notably, LP Conformer~\cite{chang2024conformer} achieved a 0.9\% WER using 100K hours of training data. More recent models, including Whisper-Flamingo~\cite{rouditchenko2024whisper} and LLaMA-AVSR~\cite{cappellazzo2024large}, leverage large-scale pretrained models (e.g., Whisper and LLMs) and have demonstrated superior performance with WERs of 0.77\% and 0.76\%, respectively. Our MMS-LLaMA surpasses previous state-of-the-art methods, achieving a WER of 0.72\% when trained on 1,759 hours of data. When trained on only 433 hours of data, MMS-LLaMA obtains a WER of 0.9\%, outperforming both Whisper-Flamingo and LLaMA-AVSR. Under noisy conditions, our model also achieves state-of-the-art performance, with a WER of 1.9\%. Notably, our model significantly improves performance in noisy conditions, addressing a major weakness of previous multimodal speech LLM framework that showed a substantial gap between noisy and clean conditions. Please note that the proposed MMS-LLaMA not only achieves the superior performances but also effectively reduces the number of tokens with the proposed AV Q-Former.

\begin{table}[t]
\renewcommand{\arraystretch}{1.3}
\renewcommand{\tabcolsep}{0.1mm}
\centering
\resizebox{0.9999\linewidth}{!}{
  \begin{tabular}{cccccc}
    \toprule
    \multirow{3}{*}{\makecell{\textbf{Method}}} 
    & \multirow{3}{*}{\makecell{\textbf{\# MMS} \\ \textbf{Tokens} \\ \textbf{per second}}}
    & \multirow{3}{*}{\makecell{\textbf{GPU} \\ \textbf{Memory} \\ \textbf{Usage(GB)}}}
    & \multirow{3}{*}{\makecell{\textbf{FLOPs} \\ \textbf{(T)}}}
    & \multirow{3}{*}{\makecell{\textbf{WER(\%)$\downarrow$}}} \\
    & & & & \\
    & & & & \\
    \hline
    \makecell{Baseline* \\ \cite{cappellazzo2024large}} & 25 & 18.2 & 2.24 & 0.97 \\
    \hdashline
    + Early AV Fusion & 12.5 & 14.7 & 1.81 & 0.92  \\
    \hdashline
    \multirow{3}{*}{+ AV Q-Former}
        & 0.826 & 12.1 & 1.35 & 1.99 \\
        & 1.818 & 12.1 & 1.39 & 1.10 \\
        & 2.822 & 12.2 & 1.42 & 0.95 \\
    \hdashline
    \multirow{3}{*}{+ Speech Rate Predictor}
        & 1.035 & 12.2 & 1.36 & 1.61 \\
        & 2.290 & 12.3 & 1.40 & 0.97 \\
        & 3.538 & 12.4 & 1.44 & 0.90 \\
    
    \bottomrule
  \end{tabular}}
\caption{Effectiveness of each proposed component on LRS3. We report token throughput, gpu memory usage, FLOPs, and WER. MultiModal Speech (MMS) tokens indicates the number of tokens derived from one second of audio-visual input that are fed into the LLM. Note that in the rows corresponding to AV Q-Former and Speech Rate Predictor, query frequencies of 1, 2, and 3 are applied from top to bottom. $^*$We re-implemented it to have the same LLM parameters with ours.}
\label{table:2}
\end{table}

\subsection{Ablation study}
\subsubsection{Validation of the Effectiveness of Each Component via Sequential Integration}
To verify the effectiveness of the proposed components in terms of computation cost and WER, we have conducted 8 experiments through sequential integration. These all models except for baseline, are applied concatenation early AV-fusion, trained using 433 hours training data. Moreover, for fair comparison, we re-implemented baseline model* \cite{cappellazzo2024large} using Llama 3.2 3B instead of LLaMA 3.1 8B, to mitigate the effects of varying LLM parameter sizes. Table~\ref{table:2} presents a detailed performance comparisons on the LRS3 dataset. 

The baseline uses 25 multimodal speech tokens and achieves a WER of 0.97\% with 2.24T FLOPs and 18.2GB gpu memory usage. With the addition of the early av fusion component, token throughput is reduced to 12.5 multimodal speech tokens per second, FLOPs decreases to 1.81T, while the WER improves slightly to 0.92\%. 

Next, we add the AV Q-Former component. Its impact is evaluated under three different query frequency configurations of 1, 2, and 3. Please note that the results using query frequencies of 1, 2, and 3 are shown from top to bottom in the table, respectively. The query frequency of 1 yields the lowest FLOPs of 1.35T and gpu memory usage of 12.1 GB, but the WER increases significantly to 1.99\%. Increasing the query frequency to 2 and 3 leads to a modest rise in FLOPs of 1.39T and 1.42T, and gpu memory usage of 12.1GB and 12.2GB, while the WER improves to 1.10\% and 0.95\%, respectively. This demonstrates a trade-off where higher query frequency counts can help recover recognition accuracy at the cost of slightly higher computational demands.

\begin{table}[t]
\renewcommand{\arraystretch}{1.2}
\renewcommand{\tabcolsep}{4mm}
\centering
\resizebox{0.9999\linewidth}{!}{
  \begin{tabular}{cccc}
    \toprule
    \multirow{2}{*}{\makecell{\textbf{Types of} \\ \textbf{AV-Fusion}}} 
    & \multirow{2}{*}{\makecell{\textbf{FLOPs(T)}}}
    & \multicolumn{2}{c}{\textbf{WER(\%)$\downarrow$}}  \\ 
    \cline{3-4}
     &  & \textbf{Noisy} & \textbf{Clean} \\
    \hline
    Concatenation & 1.50 & 2.40 & 0.90\\
    Addition  & 1.49 & 3.02 & 0.97 \\
    Multimodal Attention & 1.50 & 3.03 & 0.87  \\
    \bottomrule
  \end{tabular}}
\caption{Comparison of different audio-visual fusion strategies in terms of computational cost (FLOPs), and WER under noisy and clean conditions.}
\label{table:3}
\end{table}

Similarly, we incorporate the speech rate predictor at the same query frequency settings, to validate its effectiveness. Specifically, at a query frequency of 1, the FLOPs and GPU memory usage increases by just 0.1T and 0.1GB, leading to a WER improvement to 1.61\%. At frequencies of 2 and 3, the FLOPs slightly incrase at 1.40T and 1.44T, respectively, and GPU memory usage is nearly the same, while the WER improves further to 0.97\% and 0.9\%. Furthermore, the analysis shows that token usage increases by only 0.2, 0.4, and 0.7 tokens per second for query frequencies of 1, 2, and 3, respectively. These minor increases do not significantly affect FLOPs, yet still yield WER improvements of 0.38\%, 0.13\%, and 0.05\% under the same settings.

Overall, the sequential integration of the proposed components demonstrates that both early av fusion and the additional modules (AV Q-Former and speech rate predictor) can effectively reduce computational costs and, under appropriate configurations, maintain or improve recognition accuracy.

\subsubsection{Evaluation of Different Audio-Visual Fusion Strategies}
To reduce the computational cost in LLMs, we have introduced an early AV-fusion module that shifts the fusion process to an earlier stage. To determine the most effective approach, we conduct ablation study by using three different AV-fusion techniques: Concatenation, Addition, and Multimodal Attention.

The results, presented in Table~\ref{table:3}, indicate that all three fusion techniques exhibit similar FLOPs. Under noisy conditions, concatenation achieves the best WER of 2.4\%, while multimodal attention performs best on clean speech with 0.87\% WER, followed by concatenation of 0.9\% WER. Because the performance gap in noisy settings is more significant than in clean conditions, we adopt concatenation as our primary fusion strategy in the other experiments.

\begin{table}[t]
\centering
\resizebox{0.9999\linewidth}{!}{
  \begin{tabular}{ccccc}
    \toprule
    \multirow{2}{*}{\makecell{\textbf{Types of} \\ \textbf{LLM}}} 
    & \multirow{2}{*}{\makecell{\textbf{GPU Memory} \\ \textbf{Usage(GB)}}}
    & \multirow{2}{*}{\makecell{\textbf{Flops(T)}}}
    & \multicolumn{2}{c}{\textbf{WER $\downarrow$}}  \\ 
    \cline{4-5}
     & & & \textbf{Noisy} & \textbf{Clean} \\
     \hline
    LLaMA3.2-1B  & 9.8 & 1.19  & 3.11 & 1.11 \\
    LLaMA3.2-3B  & 12.3 & 1.50 & 2.40 & 0.90 \\
    LLaMA3.1-8B  & 16.7 & 2.17 & 2.61 & 1.02 \\
    \bottomrule
  \end{tabular}}
\caption{Comparison of gpu memory usage, FLOPs and WER based on the model size of LLMs.}
\label{table:4}

\end{table}

\subsubsection{Impact of LLM Model Size}
To investigate how the model size of LLMs affects AVSR performance, computational cost, and gpu memory usage, we conducted experiments using three models based on LLaMA with 1B, 3B, and 8B parameters. We trained each model on 433 hours of data, applied early audio visual fusion through concatenation, and incorporated an AV Q Former with a query frequency of 5, without including the speech rate predictor.

As shown in Table \ref{table:4}, the LLaMA3.2 3B model achieved the best WER of 0.9\% under clean conditions, outperforming both the LLaMA3.2 1B and LLaMA3.1 8B models. Under noisy conditions, it also achieved the lowest WER of 2.4\%. As expected, increasing the model size leads to higher GPU memory usage and computational costs. Given that the 3B model achieves the best performance with a balanced trade-off in computational cost, we use LLaMA3.2-3B for other experiments.

\begin{table}[t]
\centering
\resizebox{0.9999\linewidth}{!}{
  \begin{tabular}{cccccccc}
    \toprule
    \multirow{2}{*}{\makecell{\textbf{Query} \\ \textbf{Frequency}}} & 
    \multirow{2}{*}{\makecell{\textbf{Visual} \\ \textbf{Modality}}} & 
    \multicolumn{6}{c}{\textbf{SNR Level (dB), WER($\downarrow$)}}  \\ 
    \cline{3-8}
     & & $\infty$ & 5 & 2 & 0 & -2 & -5 \\
    \hline
    3 & - & 1.10 & 1.58 & 2.66 & 4.17 & 6.30 & 13.54 \\
    \hdashline
    1 & \checkmark & 1.99 & 2.36 & 3.15 & 4.30 & 6.62 & 12.25  \\
    2 & \checkmark & 1.10 & 1.54 & 2.13 & 3.34 & 4.72 & 9.31 \\
    3 & \checkmark & 0.95 & 1.30 & 1.83 & 2.66 & 4.16 & 7.44 \\
    4 & \checkmark & 0.91 & 1.32 & 1.84 & 2.72 & 3.95 & 7.42 \\
    5 & \checkmark & 0.92 & 1.22 & 1.74 & 2.91 & 4.26 & 7.28 \\
    \bottomrule
  \end{tabular}}
\caption{WER results at various SNR levels ($\infty$, 5, 2, 0, -2, -5 dB), where $\infty$ indicates clean audio, comparing different query frequencies with and without the visual modality.}

\label{table:5}
\end{table}

\subsubsection{Performance Across SNR Levels and Query Frequencies}
In this section, we aim to validate the effectiveness of the visual modality and evaluate AVSR performances across various query frequencies at different SNR levels. Table~\ref{table:5} presents the WER results spanning from clean settings ($\infty$ dB) to severely noisy conditions (-5 dB). 

The results indicate that incorporating the visual modality leads to a significant improvement in performance, especially in noisy environments. For example, while a query frequency of 3 without the visual modality obtain a WER of 1.10\% in clean setting and 13.54\% at -5 dB, adding the visual modality with the same query frequency reduces the WERs to 0.91\% and 7.44\%, respectively. Moreover, as the query frequency increases from 1 to 5, there is a general tendency for performance to improve at the -5 dB SNR level, aligning with the trends observed in Figure~\ref{fig:2}.

\subsection{Audio-Unavailable Scenarios}
\subsubsection{Speech Rate Predictor Ablation}
In real-world scenarios, audio signals are often distorted due to various factors. To mitigate this issue, we have trained a visual speech rate predictor that relies on lip movements.

To confirm the effectiveness of the visual speech predictor, we have replaced the audio-based speech predictor with the visual speech predictor in our AVSR framework and evaluated its performance on a pre-trained model. The results, shown in Table~\ref{table:s1}, indicate that the visual speech rate predictor achieves a WER of 0.75\%, which is nearly the same as the 0.72\% obtained by the audio-based predictor.

\begin{table}[t]
\centering
\resizebox{0.9\linewidth}{!}{
\begin{tabular}{l|c}
\hline
Method & WER (\%) \\
\hline
w/ Speech rate predictor & 0.72 \\
w/ Visual speech rate predictor & 0.75 \\
\hline
\end{tabular}}
\caption{Comparison between audio-based and visual speech rate predictors.}
\label{table:s1}
\end{table}

\subsubsection{VSR performance}
To evaluate the effectiveness of the proposed method on audio-unavailable scenarios, we have conducted an additional experiment in which the proposed method was trained solely on the visual modality. In this experiment, we have utilized a visual speech rate predictor and the LRS3 dataset for training. 

The results are illustrated in Table~\ref{table:s2}. To assess the effectiveness of our proposed approach, we compared it with a previously introduced LLM-based VSR framework (VSP-LLM) \cite{yeo2024visual}. VSP-LLM uses unit-based deduplication to reduce sequence length by up to 53\%, reporting a WER of 26.7\%. Although our method yields a slightly higher WER of 28.7\%, it's worth noting that it reduces the visual speech features per second by up to 23\%.

\subsection{Additional Results at Different SNR Levels}
To facilitate fair comparison and follow-up research, we provide additional results for our proposed method at various SNR levels. The experiments were conducted at SNR levels of -15, -10, -5, 0, 5, 10, and 15 dB, following previous works. Table~\ref{table:s3} illustrates the WERs at each SNR level. 

Across the different SNR levels, the proposed method outperforms the previous approach \cite{cheng2023mixspeech}. Specifically, performance at 10dB to 15dB is comparable to that observed in a clean environment, while substantial improvements are evident between -10dB and 0dB.

\section{Conclusion}
We have demonstrated that the proposed MMS-LLaMA, integrating efficient token compression strategies into multimodal speech LLM frameworks can dramatically reduce computational costs while preserving high-level linguistic content. By incorporating an early AV-fusion module, a dynamically adaptive AV Q-Former, and a refined query allocation strategy with a speech rate predictor, we have achieved state-of-the-art AVSR performance, attaining a WER of 0.72\% while using only 3.5 multimodal speech tokens per second. Moreover, our extensive experiments on the LRS3 dataset have confirmed that this method not only achieves a superior WER but also reduces FLOPs by 35.7\% compared to the previous LLM-based AVSR approach.

\begin{table}[t]
\centering
\resizebox{0.8\linewidth}{!}{
\begin{tabular}{l|c}
\hline
Method & WER (\%) \\
\hline
VSP-LLM \cite{yeo2024visual} & 26.7 \\
Proposed Method & 28.7 \\
\hline
\end{tabular}}
\caption{Performance comparison for the visual-only modality on LRS3.}
\label{table:s2}
\end{table}

\begin{table}[t]
\centering
\resizebox{0.9999\linewidth}{!}{
\begin{tabular}{lccccccc}
\toprule
\multirow{2}{*}{\textbf{Method}} & \multicolumn{7}{c}{\textbf{SNR (dB)}} \\
\cmidrule(lr){2-8}
& -15 & -10 & -5 & 0 & 5 & 10 & 15 \\
\midrule
\citet{cheng2023mixspeech} & -- & 34.9 & 16.6 & 5.8 & 2.6 & 2.0 & -- \\
Proposed Method & 17.1 & 13.8 & 7.0 & 1.9 & 1.0 & 0.76 & 0.75 \\
\bottomrule
\end{tabular}}
\caption{WER (\%) of different methods at various SNR levels (dB).}
\label{table:s3}
\end{table}

\section{Limitation}
While we have introduced an efficient multimodal speech LLM framework, its current focus is constrained to the AVSR task. This narrow scope may limit the immediate applicability of our framework to other domains, such as multimodal speech dialogue systems. Nonetheless, the proposed method efficiently processes audio-visual inputs by leveraging the proposed tokens reducing scheme. Building on this efficient framework, we expect that MMS-LLaMA can be extended to real-world communication scenarios by training the system on large-scale multimodal dialogue corpora.

\section*{Acknowledgments}
This work was partly supported by two funds: the National Research Foundation of Korea (NRF) grant funded by the Korea government (MSIT) (No. NRF-2022R1A2C2005529), and Institute of Information \& communications Technology Planning \& Evaluation (IITP) grant funded by the Korea government(MSIT) (No.2022-0-00124, Development of Artificial Intelligence Technology for Self-Improving Competency-Aware Learning Capabilities).

\bibliography{main}

\begin{thebibliography}{61}
\providecommand{\natexlab}[1]{#1}

\bibitem[{Afouras et~al.(2018{\natexlab{a}})Afouras, Chung, Senior, Vinyals, and Zisserman}]{afouras2018deep}
Triantafyllos Afouras, Joon~Son Chung, Andrew Senior, Oriol Vinyals, and Andrew Zisserman. 2018{\natexlab{a}}.
\newblock Deep audio-visual speech recognition.
\newblock \emph{IEEE transactions on pattern analysis and machine intelligence}, 44(12):8717--8727.

\bibitem[{Afouras et~al.(2018{\natexlab{b}})Afouras, Chung, and Zisserman}]{afouras2018lrs3}
Triantafyllos Afouras, Joon~Son Chung, and Andrew Zisserman. 2018{\natexlab{b}}.
\newblock Lrs3-ted: a large-scale dataset for visual speech recognition.
\newblock \emph{arXiv preprint arXiv:1809.00496}.

\bibitem[{Amodei et~al.(2016)Amodei, Ananthanarayanan, Anubhai, Bai, Battenberg, Case, Casper, Catanzaro, Cheng, Chen et~al.}]{amodei2016deep}
Dario Amodei, Sundaram Ananthanarayanan, Rishita Anubhai, Jingliang Bai, Eric Battenberg, Carl Case, Jared Casper, Bryan Catanzaro, Qiang Cheng, Guoliang Chen, et~al. 2016.
\newblock Deep speech 2: End-to-end speech recognition in english and mandarin.
\newblock In \emph{International conference on machine learning}, pages 173--182. PMLR.

\bibitem[{Cappellazzo et~al.(2024)Cappellazzo, Kim, Chen, Ma, Petridis, Falavigna, Brutti, and Pantic}]{cappellazzo2024large}
Umberto Cappellazzo, Minsu Kim, Honglie Chen, Pingchuan Ma, Stavros Petridis, Daniele Falavigna, Alessio Brutti, and Maja Pantic. 2024.
\newblock Large language models are strong audio-visual speech recognition learners.
\newblock \emph{arXiv preprint arXiv:2409.12319}.

\bibitem[{Chang et~al.(2024)Chang, Liao, Serdyuk, Shahy, and Siohan}]{chang2024conformer}
Oscar Chang, Hank Liao, Dmitriy Serdyuk, Ankit Shahy, and Olivier Siohan. 2024.
\newblock Conformer is all you need for visual speech recognition.
\newblock In \emph{ICASSP 2024-2024 IEEE International Conference on Acoustics, Speech and Signal Processing (ICASSP)}, pages 10136--10140. IEEE.

\bibitem[{Chen et~al.(2023)Chen, Hu, Yang, Siniscalchi, Chen, and Chng}]{chen2023hyporadise}
Chen Chen, Yuchen Hu, Chao-Han~Huck Yang, Sabato~Marco Siniscalchi, Pin-Yu Chen, and Eng-Siong Chng. 2023.
\newblock Hyporadise: An open baseline for generative speech recognition with large language models.
\newblock \emph{Advances in Neural Information Processing Systems}, 36:31665--31688.

\bibitem[{Chen et~al.(2024)Chen, Li, Hu, Siniscalchi, Chen, Chng, and Yang}]{chen2024s}
Chen Chen, Ruizhe Li, Yuchen Hu, Sabato~Marco Siniscalchi, Pin-Yu Chen, Ensiong Chng, and Chao-Han~Huck Yang. 2024.
\newblock It's never too late: Fusing acoustic information into large language models for automatic speech recognition.
\newblock \emph{arXiv preprint arXiv:2402.05457}.

\bibitem[{Cheng et~al.(2023{\natexlab{a}})Cheng, Jin, Huang, Li, Lin, Wang, Wang, Liu, Yin, and Zhao}]{cheng2023mixspeech}
Xize Cheng, Tao Jin, Rongjie Huang, Linjun Li, Wang Lin, Zehan Wang, Ye~Wang, Huadai Liu, Aoxiong Yin, and Zhou Zhao. 2023{\natexlab{a}}.
\newblock Mixspeech: Cross-modality self-learning with audio-visual stream mixup for visual speech translation and recognition.
\newblock In \emph{Proceedings of the IEEE/CVF International Conference on Computer Vision}, pages 15735--15745.

\bibitem[{Cheng et~al.(2023{\natexlab{b}})Cheng, Jin, Li, Lin, Duan, and Zhao}]{cheng2023opensr}
Xize Cheng, Tao Jin, Linjun Li, Wang Lin, Xinyu Duan, and Zhou Zhao. 2023{\natexlab{b}}.
\newblock Opensr: Open-modality speech recognition via maintaining multi-modality alignment.
\newblock \emph{arXiv preprint arXiv:2306.06410}.

\bibitem[{Chu et~al.(2023)Chu, Xu, Zhou, Yang, Zhang, Yan, Zhou, and Zhou}]{chu2023qwen}
Yunfei Chu, Jin Xu, Xiaohuan Zhou, Qian Yang, Shiliang Zhang, Zhijie Yan, Chang Zhou, and Jingren Zhou. 2023.
\newblock Qwen-audio: Advancing universal audio understanding via unified large-scale audio-language models.
\newblock \emph{arXiv preprint arXiv:2311.07919}.

\bibitem[{Chung et~al.(2018)Chung, Nagrani, and Zisserman}]{chung2018voxceleb2}
Joon~Son Chung, Arsha Nagrani, and Andrew Zisserman. 2018.
\newblock Voxceleb2: Deep speaker recognition.
\newblock \emph{arXiv preprint arXiv:1806.05622}.

\bibitem[{Chung et~al.(2017)Chung, Senior, Vinyals, and Zisserman}]{chung2017lip}
Joon~Son Chung, Andrew Senior, Oriol Vinyals, and Andrew Zisserman. 2017.
\newblock Lip reading sentences in the wild.
\newblock In \emph{2017 IEEE Conference on Computer Vision and Pattern Recognition (CVPR)}, pages 3444--3453. IEEE Computer Society.

\bibitem[{Dai et~al.(2023)Dai, Li, Li, Tiong, Zhao, Wang, Li, Fung, and Hoi}]{dai2023instructblip}
Wenliang Dai, Junnan Li, D~Li, AMH Tiong, J~Zhao, W~Wang, B~Li, P~Fung, and S~Hoi. 2023.
\newblock Instructblip: Towards general-purpose vision-language models with instruction tuning. arxiv 2023.
\newblock \emph{arXiv preprint arXiv:2305.06500}, 2.

\bibitem[{Deng et~al.(2020)Deng, Guo, Ververas, Kotsia, and Zafeiriou}]{deng2020retinaface}
Jiankang Deng, Jia Guo, Evangelos Ververas, Irene Kotsia, and Stefanos Zafeiriou. 2020.
\newblock Retinaface: Single-shot multi-level face localisation in the wild.
\newblock In \emph{Proceedings of the IEEE/CVF conference on computer vision and pattern recognition}, pages 5203--5212.

\bibitem[{Dettmers et~al.(2024)Dettmers, Pagnoni, Holtzman, and Zettlemoyer}]{dettmers2024qlora}
Tim Dettmers, Artidoro Pagnoni, Ari Holtzman, and Luke Zettlemoyer. 2024.
\newblock Qlora: Efficient finetuning of quantized llms.
\newblock \emph{Advances in Neural Information Processing Systems}, 36.

\bibitem[{Dubey et~al.(2024)Dubey, Jauhri, Pandey, Kadian, Al-Dahle, Letman, Mathur, Schelten, Yang, Fan et~al.}]{dubey2024llama}
Abhimanyu Dubey, Abhinav Jauhri, Abhinav Pandey, Abhishek Kadian, Ahmad Al-Dahle, Aiesha Letman, Akhil Mathur, Alan Schelten, Amy Yang, Angela Fan, et~al. 2024.
\newblock The llama 3 herd of models.
\newblock \emph{arXiv preprint arXiv:2407.21783}.

\bibitem[{Elman(1990)}]{elman1990finding}
Jeffrey~L Elman. 1990.
\newblock Finding structure in time.
\newblock \emph{Cognitive science}, 14(2):179--211.

\bibitem[{Fathullah et~al.(2024)Fathullah, Wu, Lakomkin, Jia, Shangguan, Li, Guo, Xiong, Mahadeokar, Kalinli et~al.}]{fathullah2024prompting}
Yassir Fathullah, Chunyang Wu, Egor Lakomkin, Junteng Jia, Yuan Shangguan, Ke~Li, Jinxi Guo, Wenhan Xiong, Jay Mahadeokar, Ozlem Kalinli, et~al. 2024.
\newblock Prompting large language models with speech recognition abilities.
\newblock In \emph{ICASSP 2024-2024 IEEE International Conference on Acoustics, Speech and Signal Processing (ICASSP)}, pages 13351--13355. IEEE.

\bibitem[{Fu et~al.(2024)Fu, Cheng, Yang, Hanting, Zhao, and Jin}]{fu2024boosting}
Dongjie Fu, Xize Cheng, Xiaoda Yang, Wang Hanting, Zhou Zhao, and Tao Jin. 2024.
\newblock Boosting speech recognition robustness to modality-distortion with contrast-augmented prompts.
\newblock In \emph{Proceedings of the 32nd ACM International Conference on Multimedia}, pages 3838--3847.

\bibitem[{Gulati et~al.(2020)Gulati, Qin, Chiu, Parmar, Zhang, Yu, Han, Wang, Zhang, Wu et~al.}]{gulati2020conformer}
Anmol Gulati, James Qin, Chung-Cheng Chiu, Niki Parmar, Yu~Zhang, Jiahui Yu, Wei Han, Shibo Wang, Zhengdong Zhang, Yonghui Wu, et~al. 2020.
\newblock Conformer: Convolution-augmented transformer for speech recognition.
\newblock \emph{arXiv preprint arXiv:2005.08100}.

\bibitem[{Haliassos et~al.(2024{\natexlab{a}})Haliassos, Mira, Chen, Landgraf, Petridis, and Pantic}]{haliassos2024unified}
Alexandros Haliassos, Rodrigo Mira, Honglie Chen, Zoe Landgraf, Stavros Petridis, and Maja Pantic. 2024{\natexlab{a}}.
\newblock Unified speech recognition: A single model for auditory, visual, and audiovisual inputs.
\newblock In \emph{The Thirty-eighth Annual Conference on Neural Information Processing Systems}.

\bibitem[{Haliassos et~al.(2024{\natexlab{b}})Haliassos, Zinonos, Mira, Petridis, and Pantic}]{haliassos2024braven}
Alexandros Haliassos, Andreas Zinonos, Rodrigo Mira, Stavros Petridis, and Maja Pantic. 2024{\natexlab{b}}.
\newblock Braven: Improving self-supervised pre-training for visual and auditory speech recognition.
\newblock In \emph{ICASSP 2024-2024 IEEE International Conference on Acoustics, Speech and Signal Processing (ICASSP)}, pages 11431--11435. IEEE.

\bibitem[{Hong et~al.(2023)Hong, Kim, Choi, and Ro}]{hong2023watch}
Joanna Hong, Minsu Kim, Jeongsoo Choi, and Yong~Man Ro. 2023.
\newblock Watch or listen: Robust audio-visual speech recognition with visual corruption modeling and reliability scoring.
\newblock In \emph{Proceedings of the IEEE/CVF Conference on Computer Vision and Pattern Recognition}, pages 18783--18794.

\bibitem[{Hong et~al.(2022)Hong, Kim, Yoo, and Ro}]{hong2022visual}
Joanna Hong, Minsu Kim, Daehun Yoo, and Yong~Man Ro. 2022.
\newblock Visual context-driven audio feature enhancement for robust end-to-end audio-visual speech recognition.
\newblock \emph{arXiv preprint arXiv:2207.06020}.

\bibitem[{Hsu et~al.(2021)Hsu, Bolte, Tsai, Lakhotia, Salakhutdinov, and Mohamed}]{hsu2021hubert}
Wei-Ning Hsu, Benjamin Bolte, Yao-Hung~Hubert Tsai, Kushal Lakhotia, Ruslan Salakhutdinov, and Abdelrahman Mohamed. 2021.
\newblock Hubert: Self-supervised speech representation learning by masked prediction of hidden units.
\newblock \emph{IEEE/ACM transactions on audio, speech, and language processing}, 29:3451--3460.

\bibitem[{Hu et~al.(2022)Hu, Hou, Chen, and Chng}]{hu2022interactive}
Yuchen Hu, Nana Hou, Chen Chen, and Eng~Siong Chng. 2022.
\newblock Interactive feature fusion for end-to-end noise-robust speech recognition.
\newblock In \emph{ICASSP 2022-2022 IEEE International Conference on Acoustics, Speech and Signal Processing (ICASSP)}, pages 6292--6296. IEEE.

\bibitem[{Huang and Kingsbury(2013)}]{huang2013audio}
Jing Huang and Brian Kingsbury. 2013.
\newblock Audio-visual deep learning for noise robust speech recognition.
\newblock In \emph{2013 IEEE international conference on acoustics, speech and signal processing}, pages 7596--7599. IEEE.

\bibitem[{Kim et~al.(2024{\natexlab{a}})Kim, Kim, and Ro}]{kim2024prompt}
Minsu Kim, Hyung-Il Kim, and Yong~Man Ro. 2024{\natexlab{a}}.
\newblock Prompt tuning of deep neural networks for speaker-adaptive visual speech recognition.
\newblock \emph{IEEE Transactions on Pattern Analysis and Machine Intelligence}.

\bibitem[{Kim et~al.(2023)Kim, Yeo, Choi, and Ro}]{kim2023lip}
Minsu Kim, Jeong~Hun Yeo, Jeongsoo Choi, and Yong~Man Ro. 2023.
\newblock Lip reading for low-resource languages by learning and combining general speech knowledge and language-specific knowledge.
\newblock In \emph{Proceedings of the IEEE/CVF International Conference on Computer Vision}, pages 15359--15371.

\bibitem[{Kim et~al.(2022)Kim, Yeo, and Ro}]{kim2022distinguishing}
Minsu Kim, Jeong~Hun Yeo, and Yong~Man Ro. 2022.
\newblock Distinguishing homophenes using multi-head visual-audio memory for lip reading.
\newblock In \emph{Proceedings of the AAAI conference on artificial intelligence}, volume~36, pages 1174--1182.

\bibitem[{Kim et~al.(2024{\natexlab{b}})Kim, Yeo, Park, Rha, and Ro}]{kim2024efficient}
Minsu Kim, Jeonghun Yeo, Se~Jin Park, Hyeongseop Rha, and Yong~Man Ro. 2024{\natexlab{b}}.
\newblock Efficient training for multilingual visual speech recognition: Pre-training with discretized visual speech representation.
\newblock In \emph{Proceedings of the 32nd ACM International Conference on Multimedia}, pages 1311--1320.

\bibitem[{Kim et~al.(2024{\natexlab{c}})Kim, Jang, Bae, Kim, and Yun}]{kim2024learning}
Sungnyun Kim, Kangwook Jang, Sangmin Bae, Hoirin Kim, and Se-Young Yun. 2024{\natexlab{c}}.
\newblock Learning video temporal dynamics with cross-modal attention for robust audio-visual speech recognition.
\newblock In \emph{2024 IEEE Spoken Language Technology Workshop (SLT)}, pages 447--454. IEEE.

\bibitem[{Kim et~al.(2017)Kim, Hori, and Watanabe}]{kim2017joint}
Suyoun Kim, Takaaki Hori, and Shinji Watanabe. 2017.
\newblock Joint ctc-attention based end-to-end speech recognition using multi-task learning.
\newblock In \emph{2017 IEEE international conference on acoustics, speech and signal processing (ICASSP)}, pages 4835--4839. IEEE.

\bibitem[{Lian et~al.(2023)Lian, Baevski, Hsu, and Auli}]{lian2023av}
Jiachen Lian, Alexei Baevski, Wei-Ning Hsu, and Michael Auli. 2023.
\newblock Av-data2vec: Self-supervised learning of audio-visual speech representations with contextualized target representations.
\newblock In \emph{2023 IEEE Automatic Speech Recognition and Understanding Workshop (ASRU)}, pages 1--8. IEEE.

\bibitem[{Ma et~al.(2023)Ma, Haliassos, Fernandez-Lopez, Chen, Petridis, and Pantic}]{ma2023auto}
Pingchuan Ma, Alexandros Haliassos, Adriana Fernandez-Lopez, Honglie Chen, Stavros Petridis, and Maja Pantic. 2023.
\newblock Auto-avsr: Audio-visual speech recognition with automatic labels.
\newblock In \emph{ICASSP 2023-2023 IEEE International Conference on Acoustics, Speech and Signal Processing (ICASSP)}, pages 1--5. IEEE.

\bibitem[{Ma et~al.(2021)Ma, Petridis, and Pantic}]{ma2021end}
Pingchuan Ma, Stavros Petridis, and Maja Pantic. 2021.
\newblock End-to-end audio-visual speech recognition with conformers.
\newblock In \emph{ICASSP 2021-2021 IEEE International Conference on Acoustics, Speech and Signal Processing (ICASSP)}, pages 7613--7617. IEEE.

\bibitem[{Ma et~al.(2022)Ma, Petridis, and Pantic}]{ma2022visual}
Pingchuan Ma, Stavros Petridis, and Maja Pantic. 2022.
\newblock Visual speech recognition for multiple languages in the wild.
\newblock \emph{Nature Machine Intelligence}, 4(11):930--939.

\bibitem[{Martinez et~al.(2020)Martinez, Ma, Petridis, and Pantic}]{martinez2020lipreading}
Brais Martinez, Pingchuan Ma, Stavros Petridis, and Maja Pantic. 2020.
\newblock Lipreading using temporal convolutional networks.
\newblock In \emph{ICASSP 2020-2020 IEEE International Conference on Acoustics, Speech and Signal Processing (ICASSP)}, pages 6319--6323. IEEE.

\bibitem[{Mroueh et~al.(2015)Mroueh, Marcheret, and Goel}]{mroueh2015deep}
Youssef Mroueh, Etienne Marcheret, and Vaibhava Goel. 2015.
\newblock Deep multimodal learning for audio-visual speech recognition.
\newblock In \emph{2015 IEEE International Conference on Acoustics, Speech and Signal Processing (ICASSP)}, pages 2130--2134. IEEE.

\bibitem[{Noda et~al.(2015)Noda, Yamaguchi, Nakadai, Okuno, and Ogata}]{noda2015audio}
Kuniaki Noda, Yuki Yamaguchi, Kazuhiro Nakadai, Hiroshi~G Okuno, and Tetsuya Ogata. 2015.
\newblock Audio-visual speech recognition using deep learning.
\newblock \emph{Applied intelligence}, 42:722--737.

\bibitem[{Petridis et~al.(2017)Petridis, Li, and Pantic}]{petridis2017end}
Stavros Petridis, Zuwei Li, and Maja Pantic. 2017.
\newblock End-to-end visual speech recognition with lstms.
\newblock In \emph{2017 IEEE international conference on acoustics, speech and signal processing (ICASSP)}, pages 2592--2596. IEEE.

\bibitem[{Petridis et~al.(2018{\natexlab{a}})Petridis, Stafylakis, Ma, Tzimiropoulos, and Pantic}]{petridis2018audio}
Stavros Petridis, Themos Stafylakis, Pingchuan Ma, Georgios Tzimiropoulos, and Maja Pantic. 2018{\natexlab{a}}.
\newblock Audio-visual speech recognition with a hybrid ctc/attention architecture.
\newblock In \emph{2018 IEEE Spoken Language Technology Workshop (SLT)}, pages 513--520. IEEE.

\bibitem[{Petridis et~al.(2018{\natexlab{b}})Petridis, Stafylakis, Ma, Cai, Tzimiropoulos, and Pantic}]{petridis2018end}
Stavros Petridis, Themos Stafylakis, Pingehuan Ma, Feipeng Cai, Georgios Tzimiropoulos, and Maja Pantic. 2018{\natexlab{b}}.
\newblock End-to-end audiovisual speech recognition.
\newblock In \emph{2018 IEEE international conference on acoustics, speech and signal processing (ICASSP)}, pages 6548--6552. IEEE.

\bibitem[{Prabhavalkar et~al.(2023)Prabhavalkar, Hori, Sainath, Schl{\"u}ter, and Watanabe}]{prabhavalkar2023end}
Rohit Prabhavalkar, Takaaki Hori, Tara~N Sainath, Ralf Schl{\"u}ter, and Shinji Watanabe. 2023.
\newblock End-to-end speech recognition: A survey.
\newblock \emph{IEEE/ACM Transactions on Audio, Speech, and Language Processing}.

\bibitem[{Radford et~al.(2023)Radford, Kim, Xu, Brockman, McLeavey, and Sutskever}]{radford2023robust}
Alec Radford, Jong~Wook Kim, Tao Xu, Greg Brockman, Christine McLeavey, and Ilya Sutskever. 2023.
\newblock Robust speech recognition via large-scale weak supervision.
\newblock In \emph{International conference on machine learning}, pages 28492--28518. PMLR.

\bibitem[{Radhakrishnan et~al.(2023)Radhakrishnan, Yang, Khan, Kumar, Kiani, Gomez-Cabrero, and Tegner}]{radhakrishnan2023whispering}
Srijith Radhakrishnan, Chao-Han~Huck Yang, Sumeer~Ahmad Khan, Rohit Kumar, Narsis~A Kiani, David Gomez-Cabrero, and Jesper~N Tegner. 2023.
\newblock Whispering llama: A cross-modal generative error correction framework for speech recognition.
\newblock \emph{arXiv preprint arXiv:2310.06434}.

\bibitem[{Rouditchenko et~al.(2024)Rouditchenko, Gong, Thomas, Karlinsky, Kuehne, Feris, and Glass}]{rouditchenko2024whisper}
Andrew Rouditchenko, Yuan Gong, Samuel Thomas, Leonid Karlinsky, Hilde Kuehne, Rogerio Feris, and James Glass. 2024.
\newblock Whisper-flamingo: Integrating visual features into whisper for audio-visual speech recognition and translation.
\newblock \emph{arXiv preprint arXiv:2406.10082}.

\bibitem[{Rubenstein et~al.(2023)Rubenstein, Asawaroengchai, Nguyen, Bapna, Borsos, Quitry, Chen, Badawy, Han, Kharitonov et~al.}]{rubenstein2023audiopalm}
Paul~K Rubenstein, Chulayuth Asawaroengchai, Duc~Dung Nguyen, Ankur Bapna, Zal{\'a}n Borsos, F{\'e}lix de~Chaumont Quitry, Peter Chen, Dalia~El Badawy, Wei Han, Eugene Kharitonov, et~al. 2023.
\newblock Audiopalm: A large language model that can speak and listen.
\newblock \emph{arXiv preprint arXiv:2306.12925}.

\bibitem[{Serdyuk et~al.(2022)Serdyuk, Braga, and Siohan}]{serdyuk2022transformer}
Dmitriy Serdyuk, Otavio Braga, and Olivier Siohan. 2022.
\newblock Transformer-based video front-ends for audio-visual speech recognition for single and multi-person video.
\newblock \emph{arXiv preprint arXiv:2201.10439}.

\bibitem[{Shi et~al.(2022{\natexlab{a}})Shi, Hsu, Lakhotia, and Mohamed}]{shi2022learning}
Bowen Shi, Wei-Ning Hsu, Kushal Lakhotia, and Abdelrahman Mohamed. 2022{\natexlab{a}}.
\newblock Learning audio-visual speech representation by masked multimodal cluster prediction.
\newblock \emph{arXiv preprint arXiv:2201.02184}.

\bibitem[{Shi et~al.(2022{\natexlab{b}})Shi, Hsu, and Mohamed}]{shi2022robust}
Bowen Shi, Wei-Ning Hsu, and Abdelrahman Mohamed. 2022{\natexlab{b}}.
\newblock Robust self-supervised audio-visual speech recognition.
\newblock \emph{arXiv preprint arXiv:2201.01763}.

\bibitem[{Stewart et~al.(2013)Stewart, Seymour, Pass, and Ming}]{stewart2013robust}
Darryl Stewart, Rowan Seymour, Adrian Pass, and Ji~Ming. 2013.
\newblock Robust audio-visual speech recognition under noisy audio-video conditions.
\newblock \emph{IEEE transactions on cybernetics}, 44(2):175--184.

\bibitem[{Tang et~al.(2023)Tang, Yu, Sun, Chen, Tan, Li, Lu, Ma, and Zhang}]{tang2023salmonn}
Changli Tang, Wenyi Yu, Guangzhi Sun, Xianzhao Chen, Tian Tan, Wei Li, Lu~Lu, Zejun Ma, and Chao Zhang. 2023.
\newblock Salmonn: Towards generic hearing abilities for large language models.
\newblock \emph{arXiv preprint arXiv:2310.13289}.

\bibitem[{Varga and Steeneken(1993)}]{varga1993assessment}
Andrew Varga and Herman~JM Steeneken. 1993.
\newblock Assessment for automatic speech recognition: Ii. noisex-92: A database and an experiment to study the effect of additive noise on speech recognition systems.
\newblock \emph{Speech communication}, 12(3):247--251.

\bibitem[{Vaswani(2017)}]{vaswani2017attention}
A~Vaswani. 2017.
\newblock Attention is all you need.
\newblock \emph{Advances in Neural Information Processing Systems}.

\bibitem[{Yeo et~al.(2024{\natexlab{a}})Yeo, Han, Kim, and Ro}]{yeo2024visual}
Jeong~Hun Yeo, Seunghee Han, Minsu Kim, and Yong~Man Ro. 2024{\natexlab{a}}.
\newblock Where visual speech meets language: Vsp-llm framework for efficient and context-aware visual speech processing.
\newblock \emph{arXiv preprint arXiv:2402.15151}.

\bibitem[{Yeo et~al.(2024{\natexlab{b}})Yeo, Kim, Kim, Rha, Han, Cheng, and Ro}]{yeo2024personalized}
Jeong~Hun Yeo, Chae~Won Kim, Hyunjun Kim, Hyeongseop Rha, Seunghee Han, Wen-Huang Cheng, and Yong~Man Ro. 2024{\natexlab{b}}.
\newblock Personalized lip reading: Adapting to your unique lip movements with vision and language.
\newblock \emph{arXiv preprint arXiv:2409.00986}.

\bibitem[{Yeo et~al.(2024{\natexlab{c}})Yeo, Kim, Choi, Kim, and Ro}]{yeo2024akvsr}
Jeong~Hun Yeo, Minsu Kim, Jeongsoo Choi, Dae~Hoe Kim, and Yong~Man Ro. 2024{\natexlab{c}}.
\newblock Akvsr: Audio knowledge empowered visual speech recognition by compressing audio knowledge of a pretrained model.
\newblock \emph{IEEE Transactions on Multimedia}.

\bibitem[{Yeo et~al.(2024{\natexlab{d}})Yeo, Kim, Watanabe, and Ro}]{yeo2024visual2}
Jeong~Hun Yeo, Minsu Kim, Shinji Watanabe, and Yong~Man Ro. 2024{\natexlab{d}}.
\newblock Visual speech recognition for languages with limited labeled data using automatic labels from whisper.
\newblock In \emph{ICASSP 2024-2024 IEEE International Conference on Acoustics, Speech and Signal Processing (ICASSP)}, pages 10471--10475. IEEE.

\bibitem[{Yu et~al.(2024)Yu, Tang, Sun, Chen, Tan, Li, Lu, Ma, and Zhang}]{yu2024connecting}
Wenyi Yu, Changli Tang, Guangzhi Sun, Xianzhao Chen, Tian Tan, Wei Li, Lu~Lu, Zejun Ma, and Chao Zhang. 2024.
\newblock Connecting speech encoder and large language model for asr.
\newblock In \emph{ICASSP 2024-2024 IEEE International Conference on Acoustics, Speech and Signal Processing (ICASSP)}, pages 12637--12641. IEEE.

\bibitem[{Zhang et~al.(2025)Zhang, Fang, Yang, and Feng}]{zhang2025llava}
Shaolei Zhang, Qingkai Fang, Zhe Yang, and Yang Feng. 2025.
\newblock Llava-mini: Efficient image and video large multimodal models with one vision token.
\newblock \emph{arXiv preprint arXiv:2501.03895}.

\end{thebibliography}

\end{document}